\documentclass[letterpaper, 10 pt, conference]{ieeeconf}  

\UseRawInputEncoding
\usepackage{graphicx}
\usepackage{float}
\usepackage{format} 

\IEEEoverridecommandlockouts                              

\title{\LARGE \bf
Trajectory and Sway Prediction Towards Fall Prevention
}

\author{Weizhuo Wang$^{1}$, Michael Raitor$^{1}$, Steve Collins$^{1}$, C. Karen Liu$^{2}$ and Monroe Kennedy III$^{1,2}$
\thanks{Authors are members of the Departments of Mechanical Engineering Departments$^{1}$ and Computer Science Departments$^{2}$, Stanford University, Stanford, CA 94305, USA. 
{\tt\small \{weizhuo2,mraitor, stevecollins,ckliu38, monroek\}@stanford.edu.} 
This work was supported by Hoffman-Yee Award from Stanford Institute for Human-Centered Artificial Intelligence (HAI). This work was supported by the Stanford Aging and Ethnogeriatrics (SAGE)
Research Center under NIH/NIA grant P30AG059307, content does not reflect official views of NIA or NIH. Youtube link: \url{https://youtu.be/I8On2oFcvhY}}
}

\begin{document}



\maketitle
\thispagestyle{empty}
\pagestyle{empty}

\begin{abstract}


Falls are the leading cause of fatal and non-fatal injuries, particularly for older persons. Imbalance can result from the body's internal causes (illness), or external causes (active or passive perturbation). 
Active perturbation results from applying an external force to a person, while passive perturbation results from human motion interacting with a static obstacle. This work proposes a metric that allows for the monitoring of the person’s torso and its correlation to active and passive perturbations. 
We show that large changes in the torso sway can be strongly correlated to active perturbations. 
We also show that we can reasonably predict the future path and expected change in torso sway by conditioning the expected path and torso sway on the past trajectory, torso motion, and the surrounding scene. 
This could have direct future applications to fall prevention. 
Results demonstrate that the torso sway is strongly correlated with perturbations. And our model is able to make use of the visual cues presented in the panorama and condition the prediction accordingly.
\end{abstract}


\section{INTRODUCTION}

Falls are the leading cause of serious injury for older persons, and with an aging world population, the mitigation of elderly falls will improve the likelihood of aging-in-place opportunities \cite{maBalanceImprovementEffects2016}. Falls are caused by a range of factors that are internal to the body such as illness or injury, as well as those that are external to the body such as an active force pushing on an individual or the individual colliding with a static obstacle \cite{rajagopalan_fall_2017}. There has been a lot of work on fall detection through methods that classify stumbling gaits \cite{hartog2021stumblemeter, aziz2012distinguishing, choi2011study, chen_reliable_2010}. As the state of the art of loss of balance prediction, the prediction horizon is very limited and only provides a binary classification label of whether a stumble has occurred or not \cite{hartog2021stumblemeter}.  And in some work, action is taken to intervene and mitigate injury when a fall is detected \cite{tamura2009wearable, chan2006human}. 

\begin{figure}[t!]
    \centering
    \vspace{-0.2cm}
    \includegraphics[width = 0.48\textwidth]{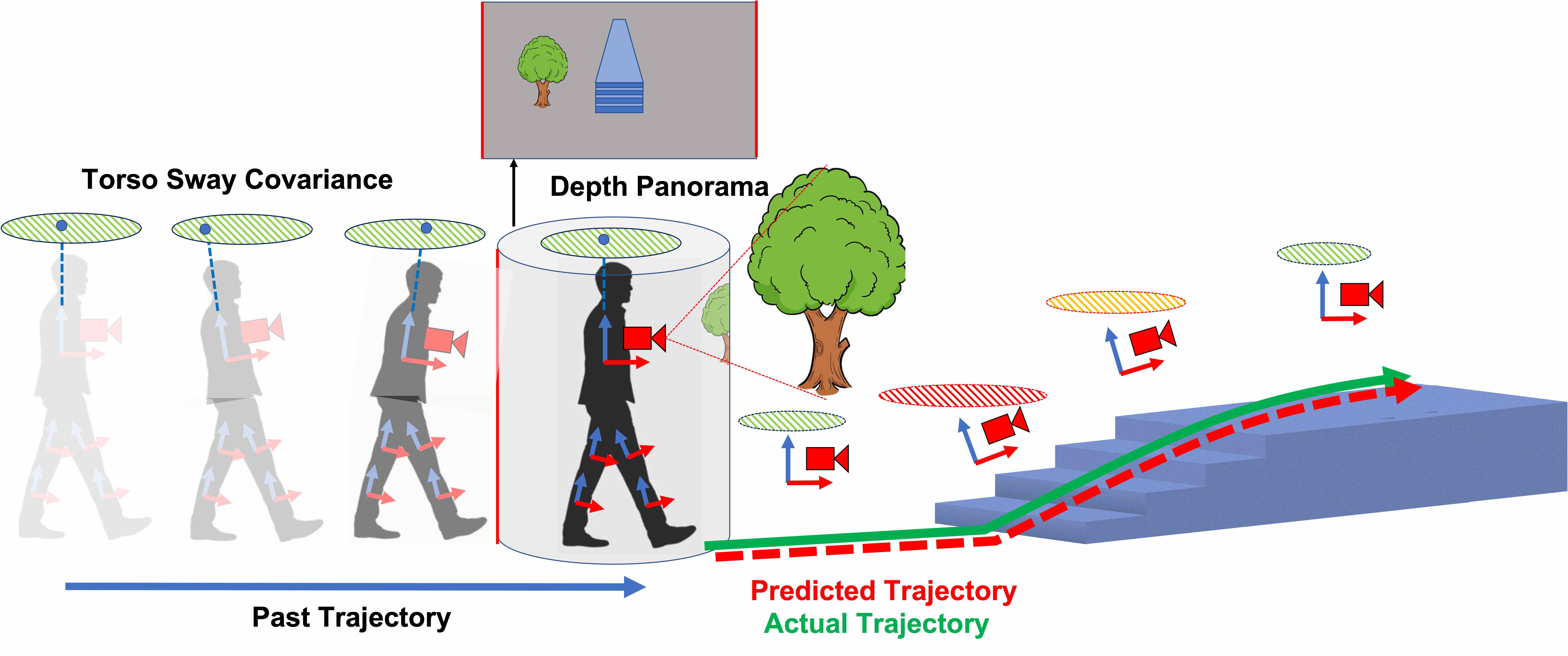}
    \vspace{-0.7cm}
    \caption{System Concept: We present a system capable of predicting future path, velocity, and torso sway conditioned on the environment and past motion of the wearer.} 
    \label{img:concept}
    \vspace{-0.6cm}
\end{figure}
But there is an important distinction between fall detection and fall prediction, if a fall can be confidently predicted then the risk to the individual can be greatly reduced, provided adequate warning is given. Fall prediction presents many inherent challenges, possibly the largest is the lack of a metric for stability that is applicable across people, physical conditions, and environments. Previous work has investigated the relationship between gait style and ultimate fall risk. In these studies, investigators have used gait observation to predict if a person is likely to fall in the coming months or identify gait features that will result in a fall due to a pre-existing illness \cite{howcroft_prospective_2017,oday_assessing_2022}. These methods utilize pattern recognition to establish a probability of a future fall but do not represent an explainable feature that is inherent to the walking motion and can be generalized. 

Classical methods of walking gait analysis leverage inverted pendulum models \cite{kuo2005energetic, hof2005condition}. Metrics that are based on the \gls*{CoP} and \gls*{CoM} can be leveraged when there are sensors capable of measuring the distribution of pressure (\gls*{CoP}) under the person's feet or the \gls*{CoM} through tracking the person's kinematic frame. Given the \gls*{CoM} and the base of support formed by the convex polytope containing feet in contact with the ground, the \gls*{MoS} can be found which measures the motion of the \gls*{CoM} relative to the support polygon. The \gls*{MoS} has been used to quantify recovery response to gait perturbations \cite{roeles2018gait}. The \gls*{CoP} motion while an individual stands or walks can be correlated with stability through definitions of `sway' path and area which are defined in the context of the \gls*{CoP} as the path the \gls*{CoP} follows and the area of triangles formed by consecutive points on the path and the average position of the \gls*{CoP} over a sliding window \cite{maBalanceImprovementEffects2016,prietoMeasuresPosturalSteadiness1996a,hufschmidtMethodsParametersBody1980,cellaDevelopmentValidationRobotic2020, linReliabilityCOPbasedPostural2008}. It is important to note that dynamic motion stability measures are not as common but very necessary as well as measurements in unstructured environments \cite{maBalanceImprovementEffects2016}. In previous work, portable wearables are used to observe and classify gait irregularity in unstructured environments through the use of \gls*{IMU} \cite{jiangElderlyFallRisk2011}. And a wide range of external and body-mounted devices have been used to observe human motion \cite{zhangRecentDevelopmentHuman2022}. When predicting an imbalance that may result from a collision with a static obstacle, it becomes necessary to predict the motion of the person in the environment. Previous work has leveraged a suite of body-mounted sensors (e.g. \gls*{IMU}) or external cameras to track the full motion of human bodies and leverage machine learning models to extrapolate expected future trajectories \cite{duBioLSTMBiomechanicallyInspired2019,sangHumanMotionPrediction2020,martinezHumanMotionPrediction2017,guoHumanMotionPrediction2019,coronaContextawareHumanMotion2020}. Longer term human motion prediction from an external observer has been used for applications of pedestrian safety for autonomous vehicles \cite{salzmannTrajectronDynamicallyFeasibleTrajectory2020}.

As far as we know, similar metrics have not been proposed to quantify loss of balance risk. We chose the torso sway vector as the metric since the torso is a main contributor to body momentum \cite{li_neuromechanical_2018}, by which stability is heavily influenced. Thus we can keep the number of sensors required as low as possible, potentially increasing the acceptability to the users, while also maintaining high sensitivity to fall risk.

Our contributions are as follows 1) We propose a metric useful for detecting both active and passive perturbations, which we've defined as torso sway covariance. 2) We present a framework for the collection of walking data in unstructured environments using a robotic wearable that is portable and physically non-invasive. Additionally, our algorithm is compatible with a neural network due to our efficient representation of environmental observations through a depth panorama image. 3) We show that we can predict the human's future trajectory and estimate the expected change in torso sway covariance given the past trajectory, past sway, and observations of the environment.
The paper is organized as follows: Sec.\ref{Method} discussed the problem formulation, system architecture, and details on how to generate panorama and sway variance ellipse. Sec.\ref{Results} reviewed the data collection setup and analyzed various results to determine the effectiveness of our approach. Finally, Sec.\ref{Conclusion} reviewed our claims and provided insights for future directions.

\section{Method} \label{Method}

\subsection{Problem Formulation}
    In this work, we develop a wearable device that actively monitors the surroundings, and predicts the future statistics of the torso motion and the expected pose of the user conditioned on past trajectories and visual features. 
    First, we establish the high correlation between disturbances and our sway covariance area metric through stage 1 treadmill data in the result section, showing that $P(\Delta\sigma_{z,t}^{peak} | perturbation_{t}) \simeq 1$. After we show the validity of our metric, we further expanded the correlation between perturbations and external features, thus allowing us to make predictions conditioned on past body motion and environment observations.
    
    For the purpose of alerting the potential fall risk, we are not trying to predict the intentional sway from the user. Because in the dataset there is no strong relation between intentionally caused sway variance and the environmental features, the model primarily learns to predict from external perturbations. Assume that the surrounding environment at  time t is captured by depth panorama $I_t \in \mathbb{R}^{H\times W}$, and the user states are described as $S_t=[p_t,v_t,\sigma_{z,t},\mathcal{G}_t,\theta_t] \in \mathbb{R}^{24}$. Given at time t: $p_t$ is position and orientation, $V_t$ is the angular and linear velocity of the torso, $\mathcal{G}_t$ is the step frequency, $I_t$ is the panorama observation of the environment, $\theta_t$ is the joint angles, and $\sigma_{z,t}$ is the sway covariance area. The goal of the VAE is to efficiently encode the information of panorama into latent space of 64 dimensions, and later expand back to full size with minimal loss such that: $z_t = VAE(I_t) \in \mathbb{R}^{64}$.
    And then the LSTM will make predictions ($\hat{\cdot{}}$ represent predicted quantities) based on the encoded observations such that: $P(\hat{S}_{t:t+m} | z_{t-n:t}, S_{z,t-n:t}), P(\hat{z}_{t:t+m} | z_{t-n:t}, S_{z,t-n:t})$.
    
    \begin{figure}[t!]
        \centering
        \includegraphics[width = 0.47\textwidth]{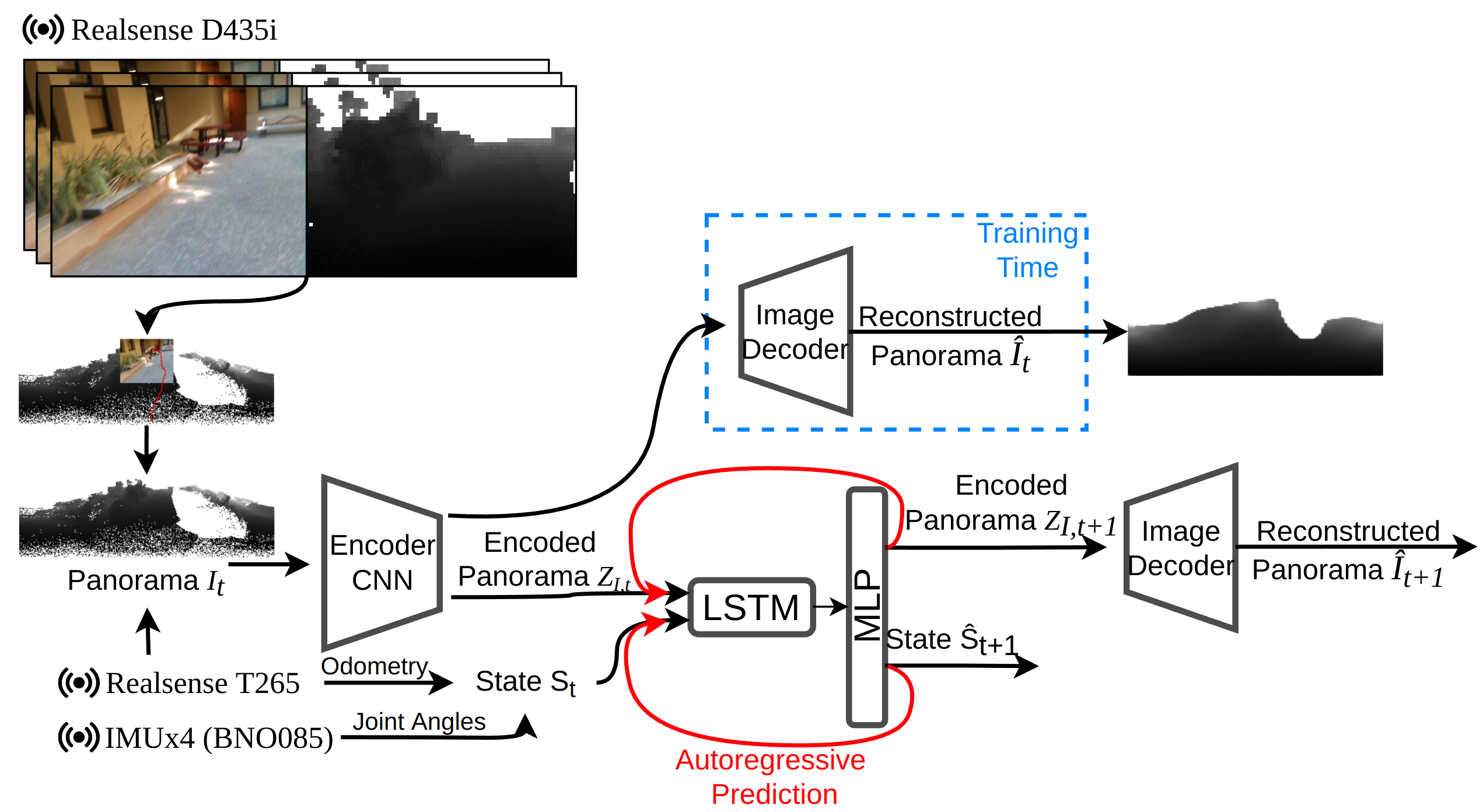}
        \vspace{-0.3cm}
        \caption{\textbf{System Architecture}. Our system is comprised of Realsense camera for vision and depth estimation and \gls*{VIO}, as well as \gls*{IMU} for the torso and each leg, thigh and calf. The network consists of a \gls*{CNN} encoder/decoder, \gls*{LSTM} (one layer), and \gls*{MLP} (two layers) for auto-regressive prediction of environment observations and future states.}
        \label{img:Arch}
        \vspace{-0.5cm}
    \end{figure}
\subsection{System Architecture}
  In order to monitor the surrounding, it is important to first determine an effective representation of the environment. Since we have little control over the scenario the user walks in, there is no fixed number of obstacles, nor do we know all the possible shapes they can pose. It is also unrealistic to build a global map or occupancy grid, as in the real-world people rarely walk in a loop. Therefore preconditioning the walking environment \cite{chakravarty_gen-slam_2019}, like how it is often done for the warehouse robot, is not applicable to our problem. We propose the use of panorama depth images which can be computed easily from the depth point cloud, takes up minimal storage space and is much more computationally efficient than directly using the point cloud or occupancy map in a neural network.

  The physical device we made consists of 4 IMUs mounted on both the left and right calf and thigh to collect accurate gait data and kinematics, along with two Intel Realsense cameras to localize the user and collect point cloud and video stream. Camera and \gls*{IMU} \acrfull{VIO} will output the global location and orientation of the torso with respect to the starting frame. The starting frame is defined as +X forward, +Y to the left, and +Z to the opposite direction of gravity. The ground plane can be estimated using a rigid body transformation based on the known mounting height of the camera. Note that the ground plane is not necessarily at the same height as the starting point.  The cameras are mounted on the torso facing forward using a plastic frame and a chest harness, while IMUs are mounted by an elastic velcro strap. Then panorama is constructed from the stream, and the sway covariance ellipse is calculated from the moving window. 
  
  A pre-processor publishes the latest states and panorama in 0.05 s intervals, or at a 20 Hz rate, meaning the sensor data streaming higher than this rate is downsampled. We don't find the need to stream any faster because the walking environment and user state change very little in merely 50 ms. We also note that many autonomous vehicles with more dynamic environments operate well with data input rates as low as 10 Hz. This avoids the hassle of processing asynchronous data rates in the network while also reducing computation load. 

\subsection{Model Architecture}
  The state and panorama are then fed into the \gls*{VAE}-\gls*{LSTM} model to predict future states. For the time series prediction, we chose LSTM for its high performance among the three most common recurrent architectures RNN, LSTM, and GRU, despite the longer training time\cite{shewalkar_performance_2019}. 
  The concept of the model is to use VAE to encode the panorama in the latent space and the LSTM will propagate the time series temporally. The result will contain predictions for all the states: position, velocity, pose, angular velocity, joint angles, step frequency and sway covariance. Further analysis can be performed on the prediction result to alert when elevated risk of falling is predicted. The resulting architecture is shown in Fig.\ref{img:Arch}. 
  
  Specifically, the encoder CNN consists of 2 layers of CNN followed by 2 layers of MLP to output 64 dimensions in latent space, the image decoder consists of 2 layers of MLP followed by 4 layers of transposed convolution layers, while the prediction model is a single LSTM layer with 2 layers of MLP at the output. The training of the whole model is separated into 2 stages. In stage 1, the panorama encoder and decoder are trained together on all the panoramas in the dataset to ensure a high-quality reconstruction. Loss is calculated as below, as given in InfoVAE paper \cite{zhao_infovae_2018} to avoid the ignored latent code problem in the reconstruction. Notice that we changed the first term to L1 loss and added weighting by pixel depth, in order to get better reconstruction close to the user and a sharper image.
  \begin{align}
      \mathcal{L} &= \beta \left(1-\frac{1}{2}\frac{I}{max(I)}\right)||\hat{I}-I||_{1}-  \nonumber\\
                    &(1-\alpha)E_{p_\mathcal{D}(x)}D_{KL}(q_{\phi}(z|x)||p(z)) - \nonumber\\ 
                    &(\alpha+\lambda-1)D_{KL}(q_{\phi}(x)||p(z))  \nonumber
  \end{align}

  In the second stage of the training we freeze the pre-trained VAE (encoder and decoder) and optimize the LSTM predictor on sub-sequences of trajectories. Specifically, we use 10 s pieces (200 time steps) with the first 7.5s as input and the last 2.5 seconds as the label. Again, the L1 loss is used for both states and panorama reconstruction:
  \vspace{-0.2cm}
  \begin{align}
      \mathcal{L}_{state} = ||\hat{S}-S||_1,
      \mathcal{L}_{Pano} = \left(1-\frac{1}{2}\frac{I}{max(I)}\right)||\hat{I}-I||_{1}  \nonumber
  \end{align}
  \vspace{-0.3cm}

    In total, there are 640K parameters. Since the network proposed is sufficiently small, our model is able to conduct inference in an average of 8.5 ms (tested on RTX 3080Ti). This equates to over 110 Hz. In practice, we believe 5-10 Hz would be sufficient for deployment, and therefore easily achievable by a majority of smartphones.


\subsection{Depth Panorama Generation}

  A key component to capturing the surrounding environment is panorama images. They are generated at a resolution of 180x360, in which each pixel represents the ray scanning a 1 deg squared box. The maximum depth is set to 10 m. To populate the panorama and provide as much information as possible to the network, the panorama constructor will maintain a fixed-length queue of collected point clouds. The panorama will be constructed from all the points collected in the queue, and rotated to the torso frame.
  
\begin{figure}[t!]
        \centering
        \includegraphics[width = 0.45\textwidth]{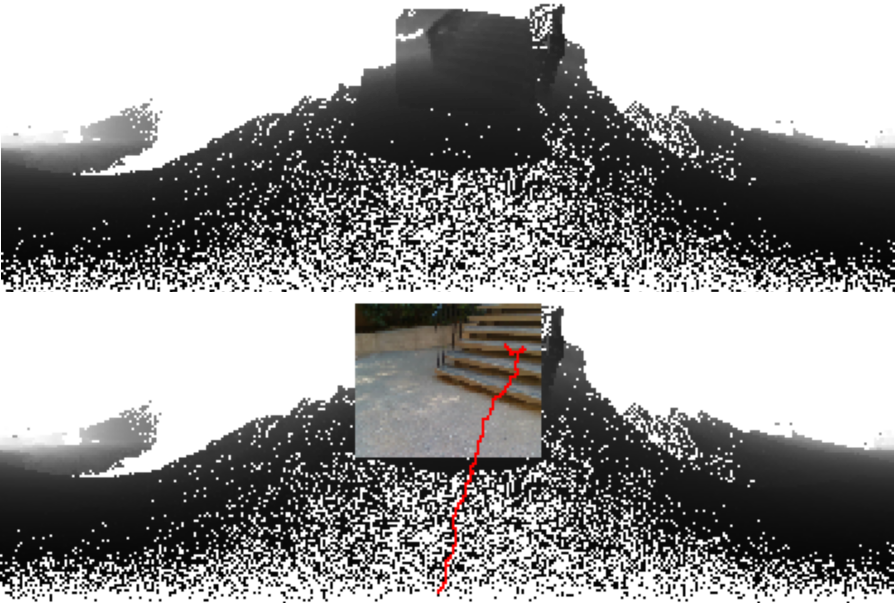}
        \vspace{-0.3cm}
        \caption{Example Panorama vs Video frame, trajectory is overlayed in red}
        \label{img:PanoConcept}
        \vspace{-0.5cm}
    \end{figure}

  An example of the resulting panorama is provided in Fig.\ref{img:PanoConcept}. Each frame only contains information about a tiny piece of the field of view. However, we can store the frame, and as we move through the environment, the point cloud that was in front of the person can move to the back of the person, thus providing additional information about the surrounding.

\subsection{Details on sway covariance}

    To quantify the risk of falling, this work proposes a new metric called Sway Covariance Ellipse Area $\sigma_z$. To calculate the sway covariance ellipse area, first, the vertical unit vector of the torso frame will be projected onto the ground plane estimated by camera \gls*{VIO}, resulting in a single point. $ Z_{proj} = R_{torso}[0,0,1]^T $ Then we will collect the projected points into a moving window, which captures the past motions of the torso. A 2D Gaussian distribution is then fitted to the moving window of points, which gives the covariance matrix and the mean for the data, therefore the name of `sway covariance'. The resulting mean vector is then the position of the ellipse, and we will use the 95\% prediction ellipse to represent the covariance matrix. The length of the major and minor axis of the fitted ellipse is the eigenvalue of the covariance matrix $cov=[a,b;b,c]$.
For 95\% of the points to fall within the ellipse, with cumulative 2D chi-squared distribution we have $\chi^2_2(5.991)=0.95$: $\text{95\% Axis $m_i$ } = \sqrt{5.991\lambda_i}$.
Then the rotation of the ellipse is defined as $\theta = atan2(\lambda_1-a,b)$.

    With the ellipse fully defined, the area of the ellipse can be easily calculated from $\sigma_z=\pi m_1 m_2$. And the change in the area can be calculated by a backward-time finite differencing, $\Delta \sigma_{z,t} = \frac{\sigma_{z,t}-\sigma_{z,t-1}}{\Delta t}$. Notice that the moving window is essentially acting as a low-pass filter, as any sudden change in sway will be considered an outlier when a Gaussian distribution is fitted. The larger the moving window is, the longer delay it takes for the algorithm to register the change and converge the fitted ellipse to the new distribution and consequently less sensitive to change in sway. In practice, we find a window of size 50, which is 2.5 s, is a good compromise between delay and sensitivity.

\section{Experiments and results} \label{Results}
We first investigated the effectiveness of using the sway Covariance ellipse compared to other metrics proposed in the previous work such as change in torso angle. We then leverage the network model trained with the proposed metric and test its performance in real-world scenarios. To have a more fine-grained loss measurement, we split the test dataset into 3 scenarios: Indoor, Outdoor Cluttered, and Outdoor Free. We also filtered the test cases based on their curvature, as predicting a straight line walking is of less interest to our goal of fall prevention. In the following sections, we will first analyze the overall performance metrics, and then compare the performance with some key cases.

    \begin{figure}[t!]
        \centering
        \includegraphics[width = 0.44\textwidth]{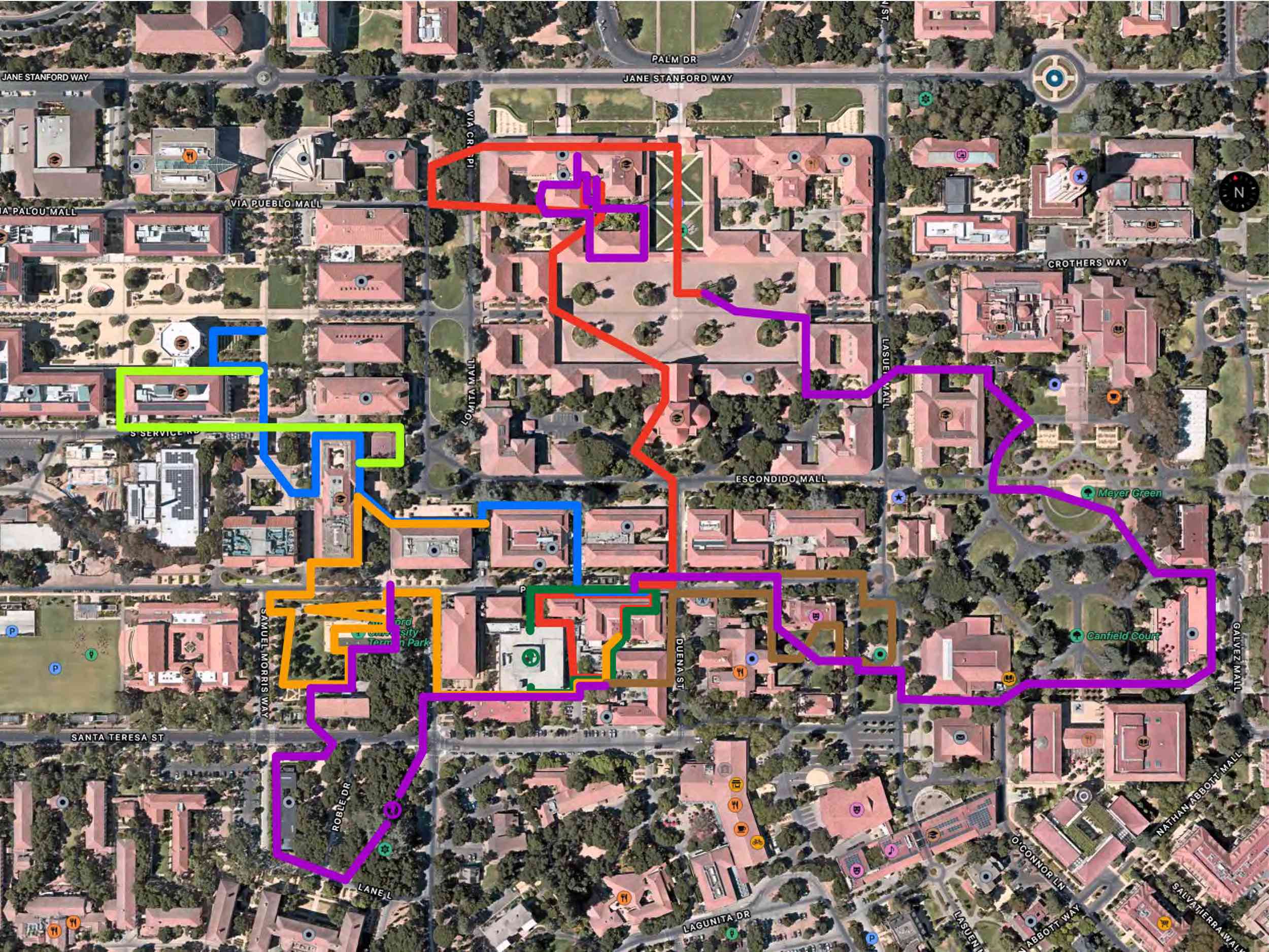}
        \vspace{-0.3cm}
        \caption{Each walking path labeled in different color}
        \label{img:S5Path}
        \vspace{-0.7cm}
    \end{figure}

\vspace{-0.2cm}
\subsection{Experiment Setup}
    The study contains two stages of data collection. Stage 1 is done on the treadmill with calibrated external perturbations exerted on pre-designed locations. While stage 2 consists of data collection when walking in the open environment with various external perturbations not pre-determined. The goal of stage 1 of the study is to build a strong correlation between external perturbations and the sway variance metric of the torso. Stage 2 builds on top of this correlation, then uses the sway metric as the middle agent to further link the perturbations to the external features. This eventually allowed us to predict the sway variance conditioning on the environmental features
    
    For stage 1 of the study, we collected data from 14 healthy participants (ages 22-31, 6 female, 8 male). Each participant walked on the treadmill at 1.25 m/s  while receiving perturbations to the pelvis. Perturbations occurred every 16-21 seconds, were in front, back, left, and right directions, and had a magnitude of 7.5\% or 15\% of body weight for a duration of 300ms. Subjects 1 and 2 each received perturbations targeted at three points in the gait cycle: heel-strike, mid-stance, and toe-off of their left foot, while subjects 3-14 only received perturbations at mid-stance. Perturbations were applied to the participants in a pseudo-random order and while they were warned that during the  walking trial, they would be bumped on the pelvis periodically, they were not given a warning before each perturbation. Participants were instructed to walk with their hands holding onto the fall-harness straps on their shoulders but were able to release the straps and move their arms freely during recovery. The absence of warning and pseudo-random ordering of perturbations blocked participant anticipatory response as much as is feasible in a laboratory experiment. Participants gave informed consent (IRB-57846). Participant motion was tracked using an optical motion tracking system (Vicon, UK) recording at 100 Hz. Participants were perturbed using an open-source perturbation system \cite{tan2020bump}. 
    
    
    
    For stage 2 of the study, 10 healthy young adults were recruited to perform real-world walking data collection. We pre-determined 10 routes across the campus that covers all kinds of common features such as a ramp, stairs, curb, unpaved surfaces, walls, doors, etc, as shown in Fig \ref{img:S5Path}. 
    Each participant was instructed to walk on two randomly picked trajectories wearing our data collection device. Participants were instructed to walk normally (with no artificial instability). Participants gave informed consent (IRB-60675). In total, we collected roughly 6000 seconds of training data. 
    \begin{figure}[t!]
        \centering
        \includegraphics[width = 0.47\textwidth]{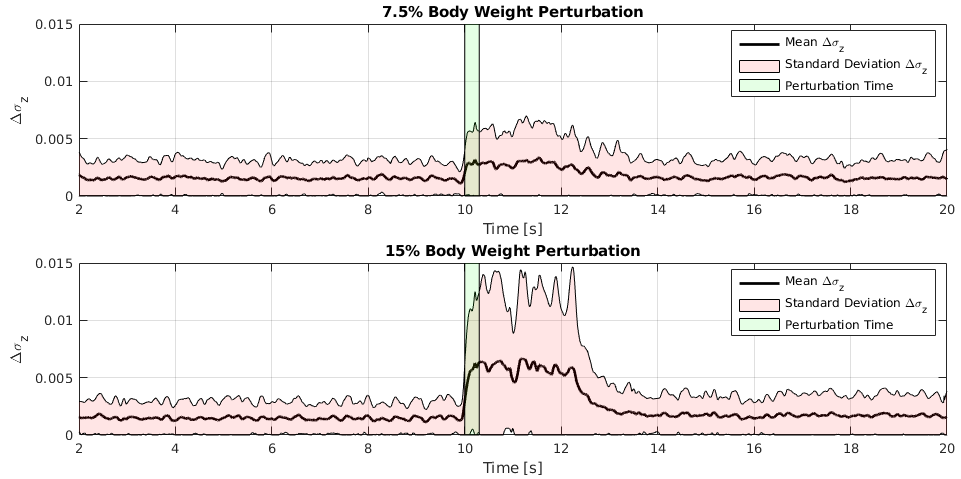}
        \vspace{-0.3cm}
        \caption{Change in Sway Covariance Ellipse Size $\Delta\sigma_z$}
        \label{img:DsigZ}
    \end{figure}
    
    \begin{figure}[t!]
        \vspace{-0.3cm}
        \centering
        \includegraphics[width = 0.47\textwidth]{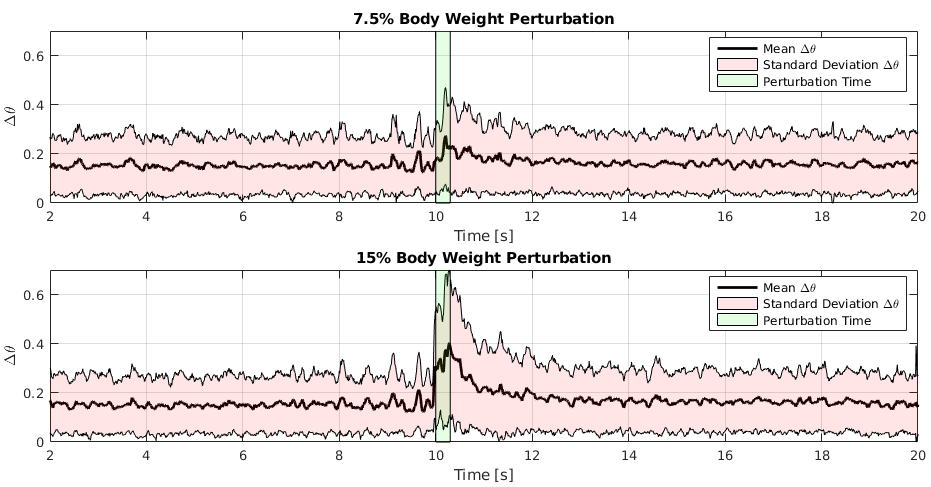}
        \vspace{-0.3cm}
        \caption{Change in Torso Angle $\Delta\theta$}
        \label{img:Dtheta}
        \vspace{-0.6cm}
    \end{figure}

\vspace{-0.1cm}
\subsection{Perturbation Detection using Sway Covariance Ellipse}
    We were able to apply the Sway Covariance Ellipse metric on the stage 1 data. There is a total of 192 walking perturbation trials from 14 subjects and force magnitude combination. The mean is shown in Figure \ref{img:DsigZ} in a thick black line and the standard deviation in pale red. We observe a significant change in the ellipse size immediately following the perturbation, and the change in size lasted throughout the recovery period. For the 7.5\% perturbation trials, the mean change of ellipse size has jumped on average 200\%, while for the 15\% perturbation trials, it jumped over 400\%. These significant changes make the sway Covariance ellipse a reliable and effective metric for perturbation detection. We would also like to emphasize that, our method does not rely on intrusive/complicated sensor setup on the subject but instead, only a torso mounted \gls*{IMU}. We do recognize that the arms play a heavy role in the recovery process from perturbations. However, since we are only quantifying the sway perturbation, not studying the exact mechanism of the recovery, modeling the torso without the extremities is sufficient for our purpose.

\subsection{Perturbation Detection using Alternative Method}
    Some may argue that the change in torso angle with respect to the ground norm ($\Delta \theta_z$) can also be a good candidate metric. Here we present the result of such metric in Figure \ref{img:Dtheta}. One quick observation is that, although the change in torso angle did show a peak, it quickly diminishes back to the noise level and becomes hardly noticeable. Further comparison of the peak height shows, for 7.5\% perturbation trials, the peak is 170\% the magnitude of noise, and for 15\% perturbation trials, the peak is 250\% the magnitude of noise. The performance is clearly inferior to our method when considering future applications of classification. In fact, we would argue that the change in the torso angle is already captured in our method as the torso angle is encoded in $Z_{proj}$ and is a part of the fitted distribution. Since our method also looks at the past history of the torso angles on different axes, better performance is expected.
    

\subsection{State Prediction in different scenarios}

\begin{figure}[t!]
    \centering
    \includegraphics[width = 0.48\textwidth]{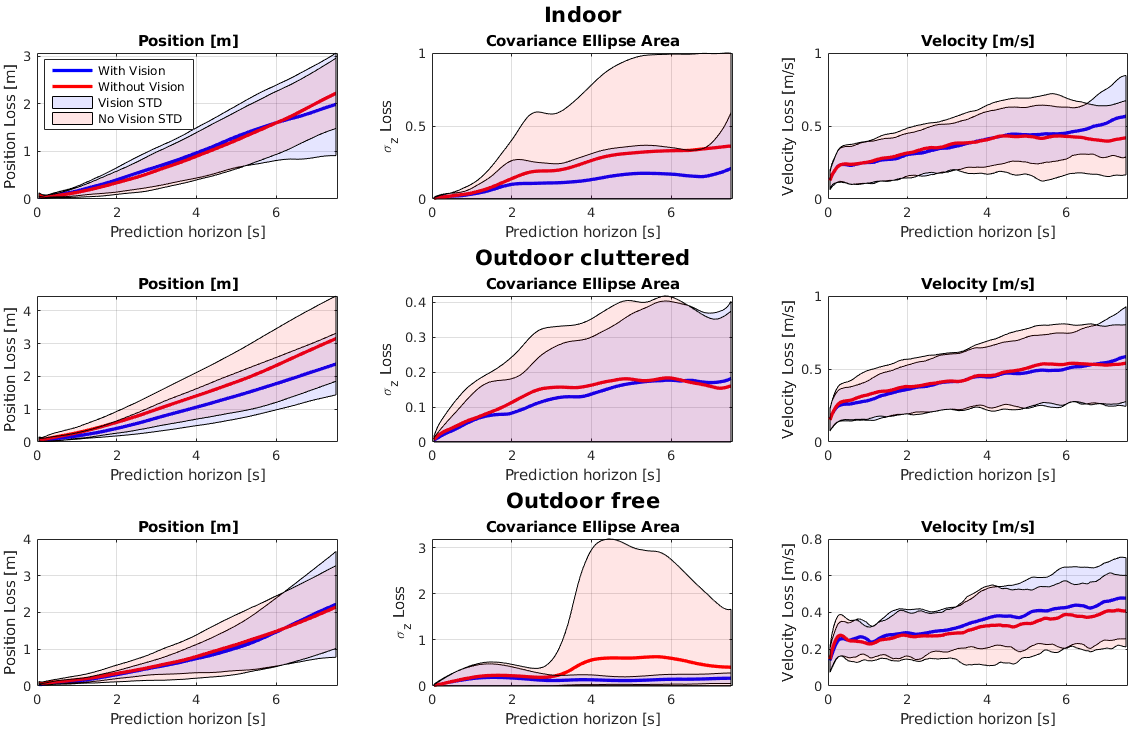}
    \vspace{-0.6cm}
    \caption{Long term loss on cases filtered by path curvature}
    \label{img:LongtermLoss9}
    \vspace{-0.6cm}
\end{figure}

    The test trajectory is divided into 3 scenarios: Indoor, Outdoor Cluttered, and Outdoor Free. Given the nature of walking, the majority of the time the participants were walking in straight lines, making it of less interest to the path prediction problem. We filtered the test set by the curvature of the path to ones with minimum turning radius $<$ 2 meters. The position and velocity losses are calculated by the spatial temporal vector distance. The Sway Covariance Ellipse Area loss is simply an L1 loss between the ground truth ellipse area and the predicted ellipse area. The results are compiled in Fig.\ref{img:LongtermLoss9}. As an ablation, we also trained the same model without vision, skipping the VAE and sending zero vector as an encoded panorama to the LSTM. The loss progression over the whole prediction horizon is shown in Fig.\ref{img:LongtermLoss9}.

    We found a comparable velocity prediction performance in all scenarios between the vision and no-vision models. However, the vision model performed better in the Sway Covariance Area prediction. The mean error is cut in half and the prediction by the vision model is far more stable. We believe the huge jump in ellipse area standard deviation for the no-vision model is caused by the accumulated error. Towards the end of the prediction horizon, the error is so large that the model is basically randomly guessing.
\begin{figure}[t!]
\centering
\includegraphics[width = 0.43\textwidth]{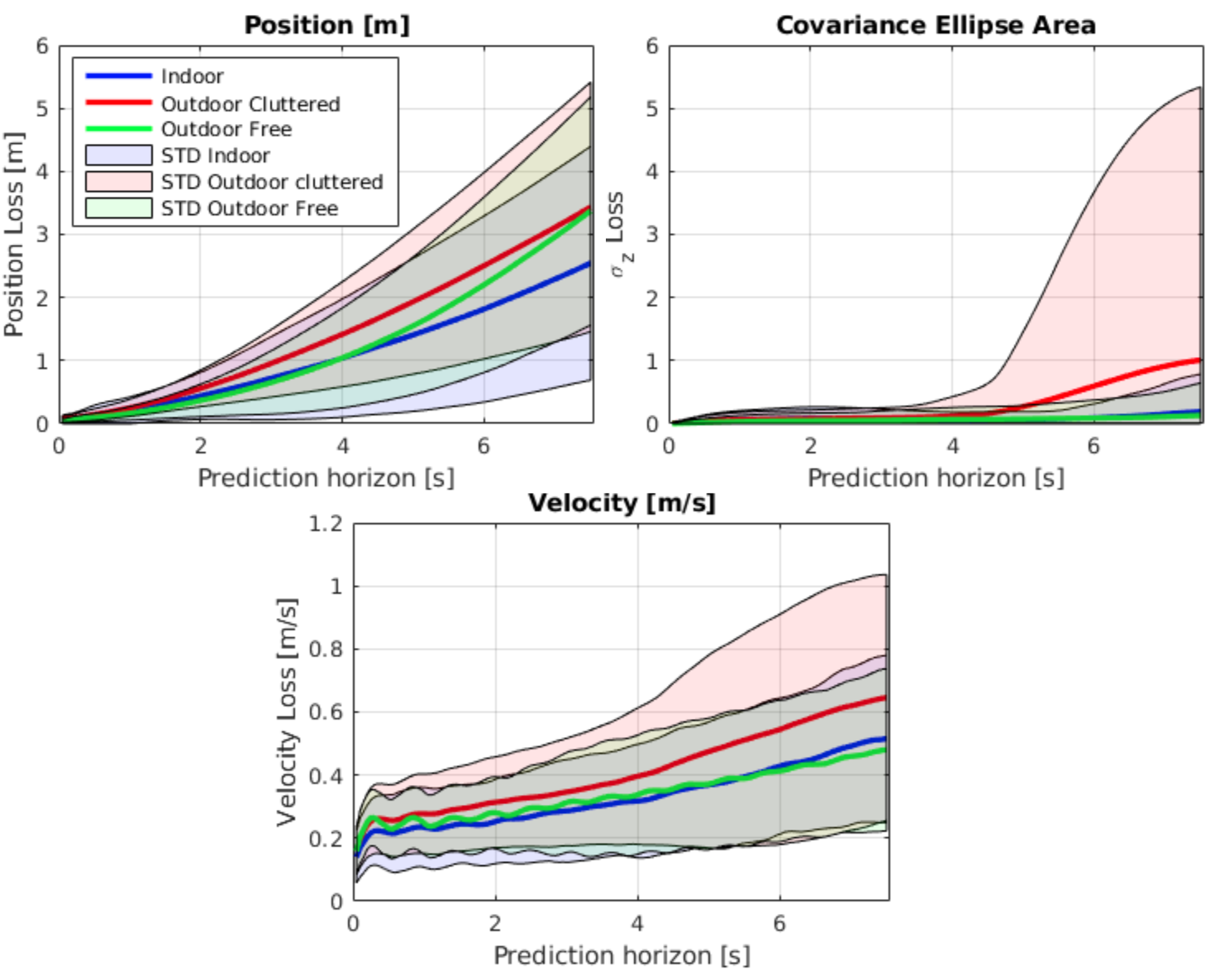}
\vspace{-0.4cm}
\caption{Long term loss on all cases}
\label{img:LongtermLoss3}
\vspace{-0.7cm}
\end{figure}

    For the position prediction, the vision model is doing well in the Outdoor Cluttered scene in which there are grounded obstacles but no walls. The mean and STD of loss of the vision model is roughly 30\% lower over the whole prediction horizon. In the outdoor free scenario, although the mean error is close, the vision model again shows a much more stable prediction over the first 2/3 of the prediction horizon. This result shows that our model is able to pick up visual cues and make effective predictions. An interesting thing to note is that: in all cases, we observe a significant deterioration in performance after 6 s for the vision model. Given our panorama input has a maximum depth of 10 m, meaning that in roughly 6 seconds, the user would be walking right up to the edge of the panorama where the model has little visual information to use. At this moment, it makes sense that the vision model will degrade back to or have worse performance compared to the no-vision model.
    
    Now we focus on the vision model and compare the result in different scenarios (Fig.\ref{img:LongtermLoss3}). Our model performs better in the Indoor scenario as the blue lines are the lowest. In the Outdoor Free scenario, the model was doing great initially, but the position prediction deteriorated quickly. This is rather expected as there are numerous equally possible paths in the free environment, and little obstacles to cut down the possibilities. The Outdoor Cluttered scenario remains the toughest scenario, and quite often the model prediction of Sway Covariance explodes after 4 s.
        
    \begin{figure}[t]
        \centering
        \includegraphics[width = 0.44\textwidth]{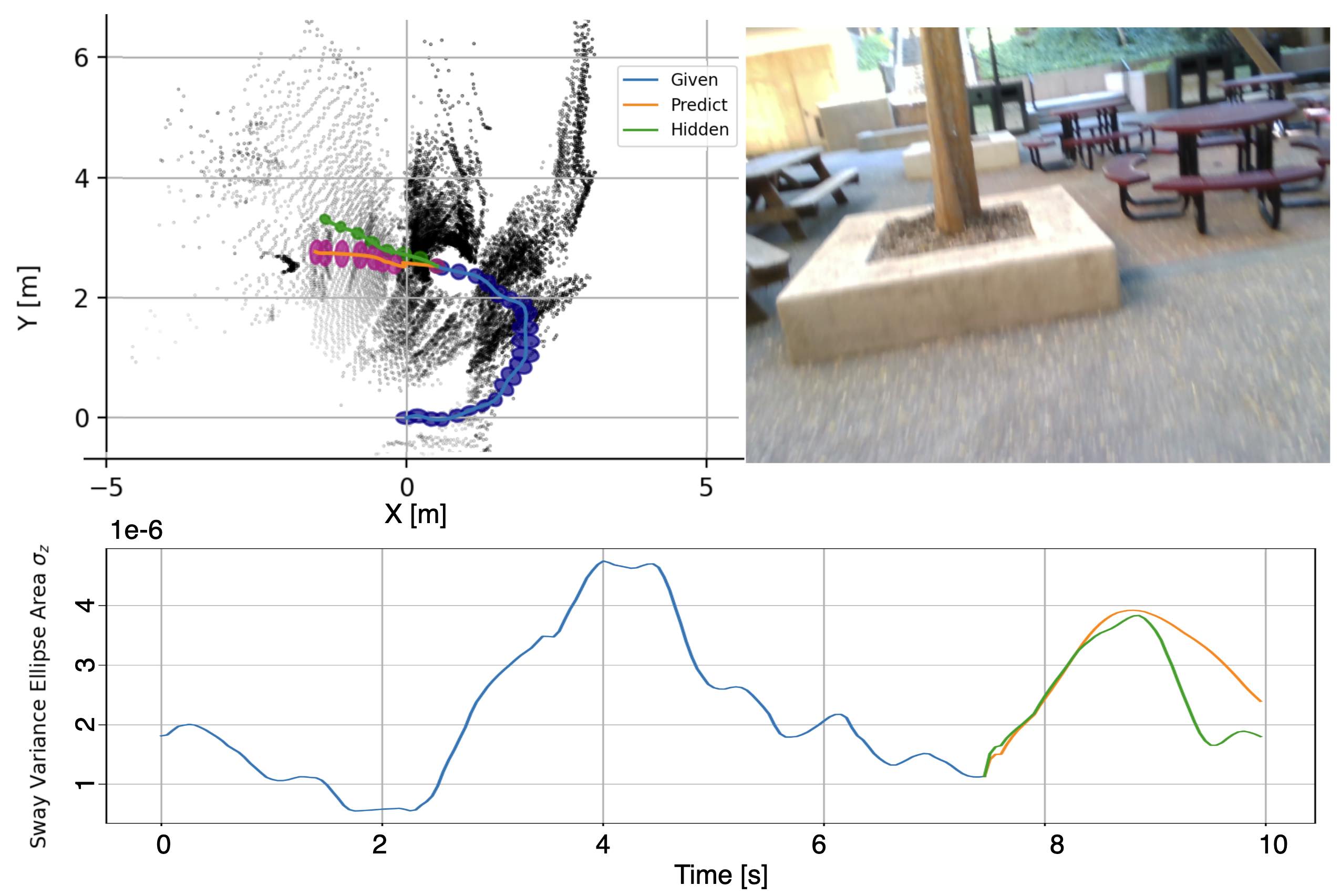}
        \vspace{-0.3cm}
        \caption{Example Ellipse Predictions Scenario 1}
        \label{img:S5KeyE100}
        \vspace{-0.3cm}
    \end{figure}

    \begin{figure}[t!]
        \centering
        \includegraphics[width = 0.44\textwidth]{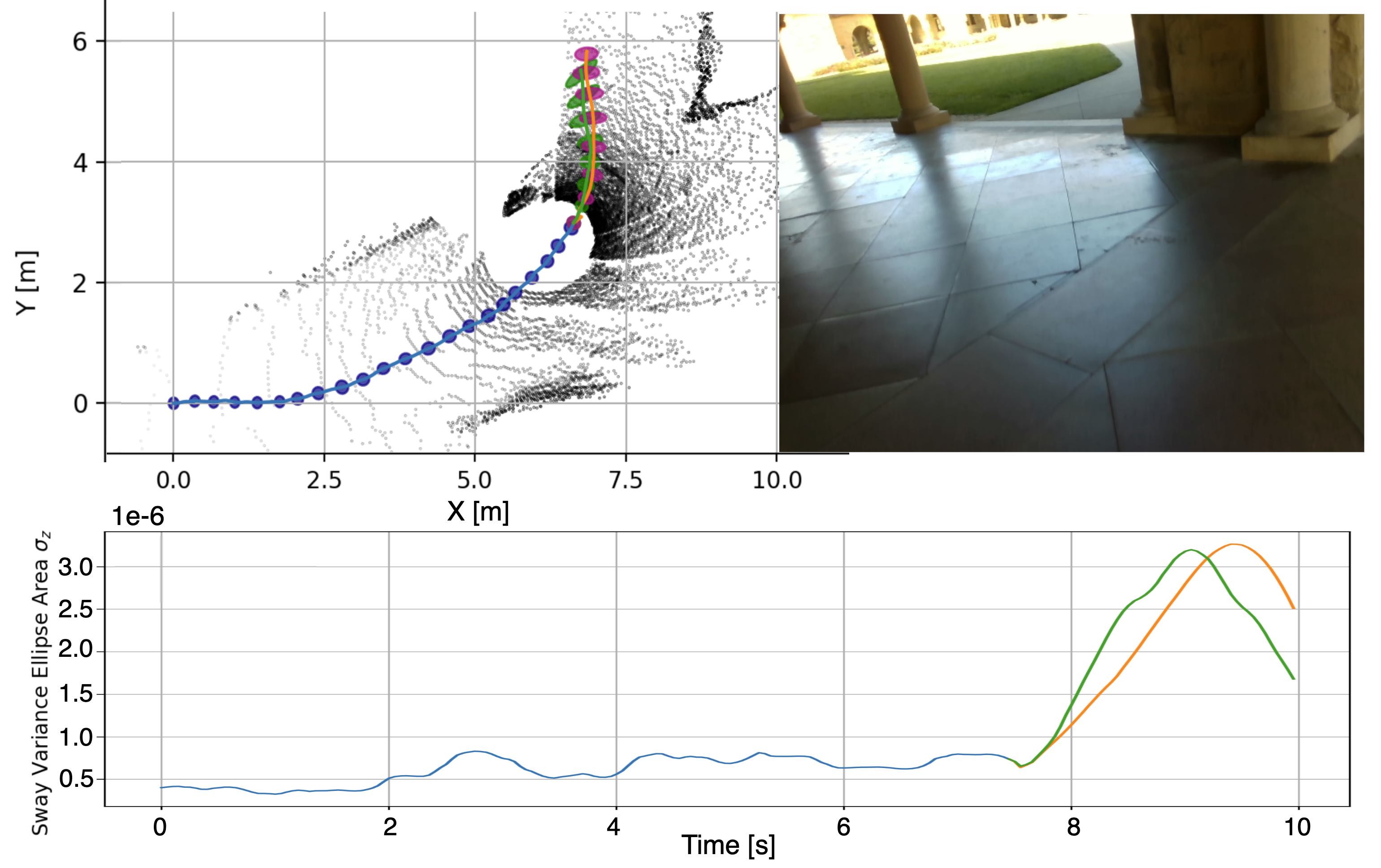}
        \vspace{-0.3cm}
        \caption{Example Ellipse Predictions Scenario 2}
        \label{img:S5KeyE5450}
        \vspace{-0.6cm}
    \end{figure}

\subsection{Prediction compared between vision and no-vision model}
    
\begin{figure}[t!]
        \centering
        \includegraphics[width = 0.48\textwidth]{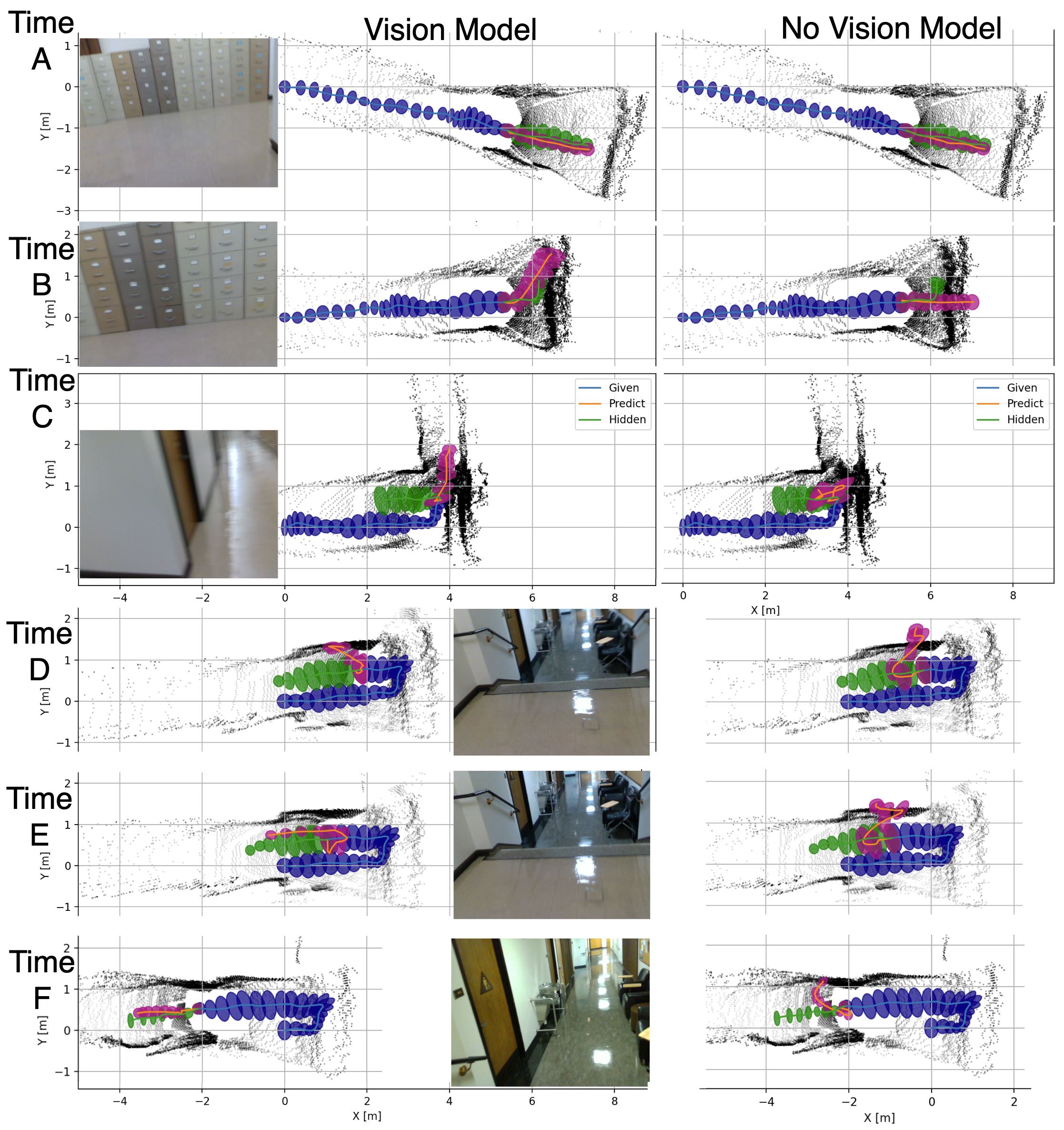}
        \vspace{-0.4cm}
        \caption{Indoor Prediction comparison between no-vision and vision model. Given trajectory is in blue, ground truth is in green, and predicted in purple. The ellipses are the corresponding sway covariance at that time instance. Black dots are the projected point cloud down to the ground plane, therefore walls and obstacles can be easily identified by dark lines.} 
        \label{img:S5KeyPred}
        \vspace{-0.67cm}
    \end{figure}

    The loss curves tell part of the story, but not the full story. It is important to watch how the two models behave in the real world. We fully recognize the limitation of our current loss function that it discourages a multimodal prediction. And unfortunately, it is quite often there exist many valid future paths in the scene, while the loss function only rewards one of the paths. We have picked an indoor example in Fig.\ref{img:S5KeyPred} to convey this. Initially, at time A, both models have no problem predicting straight lines. But as soon as there are obstacles in front, the difference is starting to appear. At time B, the vision model predicts a left turn, avoiding the wall of boxes directly in the walking path. While the blind model has no clue what is in front and kept predicting a straight line. The blind model only begins to react after the user initiated a turn. Since the training data could potentially turn in any direction, it found the best strategy is to try all directions to minimize the loss. Meanwhile, the vision model is confirming its choice of the left turn, which is completely valid. Even a human observer would have a hard time predicting the 180-degree turn. At time D, the vision model starts to correct for the 180 deg turn. Notice the predicted path never went through the wall but terminated on the wall or turned away, while the no-vision model have no idea where the wall is. This clearly shows that the vision model is picking up the visual cues as designed and is using them to improve the prediction. The time E and F show the correction process, the vision model quickly follows the correct heading, while the no-vision model continues to be confused. This also explains the lower standard deviation in the position prediction generated by the vision model.

\subsection{Sway Ellipse Area Prediction}
    We also picked several cases with elevated sway covariance, signaling potential risk of falling. In the first case (Fig.\ref{img:S5KeyE100}), the user is avoiding the tree, through the narrow openings between the concrete block and the chairs. Our model is able to predict the intended path, along with a peak of sway covariance with high accuracy. With the provided state history, there is no cue that the sway covariance will go up, ruling out the model is simply pattern finding. This is a strong indication that our model is able to pick up grounded obstacles and adjust the predicted states accordingly. In the second case (Fig.\ref{img:S5KeyE5450}), the user is making a left turn in a relatively free outdoor corridor. The pillars are shown as a dark line. Our vision model is again able to accurately predict the sudden sharp turn and the associated increase in sway covariance, despite no grounded obstacle along the way.

\section{Conclusion} \label{Conclusion}
    In this work, we proposed a metric of \textit{change in torso sway} that is strongly correlated with active and passive perturbations. For prediction, we trained a \gls*{VAE}-\gls*{LSTM} model and demonstrated the ability to predict future trajectories and torso sway conditioned on past motion and observations. We showed that our model is able to pick up visual cues from the vision input represented efficiently by depth panorama images. All of these are done on real-world data. Future directions are to develop a method to determine the best time and strategy to alert the user of potential increased fall risk. To address limitations, future steps will leverage models capable of predicting multi-hypotheses of future motion and leverage semantic labeling for better scene understanding, as well as deploy on a mobile device.









\clearpage
\bibliographystyle{IEEEtran}
\bibliography{Bibs/wang_bib, Bibs/kennedy_bib, Bibs/raitor_bib}

\begin{thebibliography}{10}
\providecommand{\url}[1]{#1}
\csname url@samestyle\endcsname
\providecommand{\newblock}{\relax}
\providecommand{\bibinfo}[2]{#2}
\providecommand{\BIBentrySTDinterwordspacing}{\spaceskip=0pt\relax}
\providecommand{\BIBentryALTinterwordstretchfactor}{4}
\providecommand{\BIBentryALTinterwordspacing}{\spaceskip=\fontdimen2\font plus
\BIBentryALTinterwordstretchfactor\fontdimen3\font minus
  \fontdimen4\font\relax}
\providecommand{\BIBforeignlanguage}[2]{{%
\expandafter\ifx\csname l@#1\endcsname\relax
\typeout{** WARNING: IEEEtran.bst: No hyphenation pattern has been}%
\typeout{** loaded for the language `#1'. Using the pattern for}%
\typeout{** the default language instead.}%
\else
\language=\csname l@#1\endcsname
\fi
#2}}
\providecommand{\BIBdecl}{\relax}
\BIBdecl

\bibitem{maBalanceImprovementEffects2016}
C.~Z.-H. Ma, D.~W.-C. Wong, W.~K. Lam, A.~H.-P. Wan, and W.~C.-C. Lee,
  ``Balance {{Improvement Effects}} of {{Biofeedback Systems}} with
  {{State-of-the-Art Wearable Sensors}}: {{A Systematic Review}},''
  \emph{Sensors}, vol.~16, no.~4, p. 434, Apr. 2016.

\bibitem{rajagopalan_fall_2017}
\BIBentryALTinterwordspacing
R.~Rajagopalan, I.~Litvan, and T.-P. Jung, ``\BIBforeignlanguage{en}{Fall
  {Prediction} and {Prevention} {Systems}: {Recent} {Trends}, {Challenges}, and
  {Future} {Research} {Directions}},'' \emph{\BIBforeignlanguage{en}{Sensors}},
  vol.~17, no.~11, p. 2509, Nov. 2017, number: 11 Publisher: Multidisciplinary
  Digital Publishing Institute. [Online]. Available:
  \url{https://www.mdpi.com/1424-8220/17/11/2509}
\BIBentrySTDinterwordspacing

\bibitem{hartog2021stumblemeter}
D.~d. Hartog, J.~Harlaar, and G.~Smit, ``The stumblemeter: design and
  validation of a system that detects and classifies stumbles during gait,''
  \emph{Sensors}, vol.~21, no.~19, p. 6636, 2021.

\bibitem{aziz2012distinguishing}
O.~Aziz, E.~J. Park, G.~Mori, and S.~N. Robinovitch, ``Distinguishing
  near-falls from daily activities with wearable accelerometers and gyroscopes
  using support vector machines,'' in \emph{2012 Annual International
  Conference of the IEEE Engineering in Medicine and Biology Society}.\hskip
  1em plus 0.5em minus 0.4em\relax IEEE, 2012, pp. 5837--5840.

\bibitem{choi2011study}
Y.~Choi, A.~Ralhan, and S.~Ko, ``A study on machine learning algorithms for
  fall detection and movement classification,'' in \emph{2011 International
  Conference on Information Science and Applications}.\hskip 1em plus 0.5em
  minus 0.4em\relax IEEE, 2011, pp. 1--8.

\bibitem{chen_reliable_2010}
G.-C. Chen, C.-N. Huang, C.-Y. Chiang, C.-J. Hsieh, and C.-T. Chan,
  ``\BIBforeignlanguage{en}{A {Reliable} {Fall} {Detection} {System} {Based} on
  {Wearable} {Sensor} and {Signal} {Magnitude} {Area} for {Elderly}
  {Residents}},'' in \emph{\BIBforeignlanguage{en}{Aging {Friendly}
  {Technology} for {Health} and {Independence}}}, ser. Lecture {Notes} in
  {Computer} {Science}, Y.~Lee, Z.~Z. Bien, M.~Mokhtari, J.~T. Kim, M.~Park,
  J.~Kim, H.~Lee, and I.~Khalil, Eds.\hskip 1em plus 0.5em minus 0.4em\relax
  Berlin, Heidelberg: Springer, 2010, pp. 267--270.

\bibitem{tamura2009wearable}
T.~Tamura, T.~Yoshimura, M.~Sekine, M.~Uchida, and O.~Tanaka, ``A wearable
  airbag to prevent fall injuries,'' \emph{IEEE Transactions on Information
  Technology in Biomedicine}, vol.~13, no.~6, pp. 910--914, 2009.

\bibitem{chan2006human}
C.~S. Chan, G.~Shi, Y.~Luo, G.~Zhang, W.~J. Li, P.~H. Leong, and K.-S. Leung,
  ``A human-airbag system for hip protection using mems motion sensors:
  Experimental feasibility results,'' in \emph{2006 International Conference on
  Mechatronics and Automation}.\hskip 1em plus 0.5em minus 0.4em\relax IEEE,
  2006, pp. 831--836.

\bibitem{howcroft_prospective_2017}
\BIBentryALTinterwordspacing
J.~Howcroft, J.~Kofman, and E.~D. Lemaire,
  ``\BIBforeignlanguage{en}{Prospective {Fall}-{Risk} {Prediction} {Models} for
  {Older} {Adults} {Based} on {Wearable} {Sensors}},''
  \emph{\BIBforeignlanguage{en}{IEEE Transactions on Neural Systems and
  Rehabilitation Engineering}}, vol.~25, no.~10, pp. 1812--1820, Oct. 2017.
  [Online]. Available: \url{https://ieeexplore.ieee.org/document/7886263/}
\BIBentrySTDinterwordspacing

\bibitem{oday_assessing_2022}
\BIBentryALTinterwordspacing
J.~O'Day, M.~Lee, K.~Seagers, S.~Hoffman, A.~Jih-Schiff, Å.~Kidziński,
  S.~Delp, and H.~Bronte-Stewart, ``Assessing inertial measurement unit
  locations for freezing of gait detection and patient preference,''
  \emph{Journal of NeuroEngineering and Rehabilitation}, vol.~19, no.~1, p.~20,
  Feb. 2022. [Online]. Available:
  \url{https://doi.org/10.1186/s12984-022-00992-x}
\BIBentrySTDinterwordspacing

\bibitem{kuo2005energetic}
A.~D. Kuo, J.~M. Donelan, and A.~Ruina, ``Energetic consequences of walking
  like an inverted pendulum: step-to-step transitions,'' \emph{Exercise and
  sport sciences reviews}, vol.~33, no.~2, pp. 88--97, 2005.

\bibitem{hof2005condition}
A.~Hof, M.~Gazendam, and W.~Sinke, ``The condition for dynamic stability,''
  \emph{Journal of biomechanics}, vol.~38, no.~1, pp. 1--8, 2005.

\bibitem{roeles2018gait}
S.~Roeles, P.~Rowe, S.~Bruijn, C.~Childs, G.~Tarfali, F.~Steenbrink, and
  M.~Pijnappels, ``Gait stability in response to platform, belt, and sensory
  perturbations in young and older adults,'' \emph{Medical \& biological
  engineering \& computing}, vol.~56, no.~12, pp. 2325--2335, 2018.

\bibitem{prietoMeasuresPosturalSteadiness1996a}
T.~Prieto, J.~Myklebust, R.~Hoffmann, E.~Lovett, and B.~Myklebust, ``Measures
  of postural steadiness: Differences between healthy young and elderly
  adults,'' \emph{IEEE Transactions on Biomedical Engineering}, vol.~43, no.~9,
  pp. 956--966, Sep. 1996.

\bibitem{hufschmidtMethodsParametersBody1980}
A.~Hufschmidt, J.~Dichgans, K.~H. Mauritz, and M.~Hufschmidt, ``Some methods
  and parameters of body sway quantification and their neurological
  applications,'' \emph{Archiv f\"ur Psychiatrie und Nervenkrankheiten}, vol.
  228, no.~2, pp. 135--150, May 1980.

\bibitem{cellaDevelopmentValidationRobotic2020}
A.~Cella, A.~D. Luca, V.~Squeri, S.~Parodi, F.~Vallone, A.~Giorgeschi,
  B.~Senesi, E.~Zigoura, K.~L.~Q. Guerrero, G.~Siri, L.~D. Michieli, J.~Saglia,
  C.~Sanfilippo, and A.~Pilotto, ``Development and validation of a robotic
  multifactorial fall-risk predictive model: {{A}} one-year prospective study
  in community-dwelling older adults,'' \emph{PLOS ONE}, vol.~15, no.~6, p.
  e0234904, Jun. 2020.

\bibitem{linReliabilityCOPbasedPostural2008}
D.~Lin, H.~Seol, M.~A. Nussbaum, and M.~L. Madigan, ``Reliability of
  {{COP-based}} postural sway measures and age-related differences,''
  \emph{Gait \& Posture}, vol.~28, no.~2, pp. 337--342, Aug. 2008.

\bibitem{jiangElderlyFallRisk2011}
S.~Jiang, B.~Zhang, and D.~Wei, ``The {{Elderly Fall Risk Assessment}} and
  {{Prediction Based}} on {{Gait Analysis}},'' in \emph{2011 {{IEEE}} 11th
  {{International Conference}} on {{Computer}} and {{Information Technology}}},
  Aug. 2011, pp. 176--180.

\bibitem{zhangRecentDevelopmentHuman2022}
J.~Zhang and M.~D. Kennedy, ``Recent {{Development}} in {{Human Motion}} and
  {{Gait Prediction}},'' in \emph{Robotics {{Retrospectives}} - {{Workshop}} at
  {{RSS}} 2020}, Jul. 2022.

\bibitem{duBioLSTMBiomechanicallyInspired2019}
X.~Du, R.~Vasudevan, and M.~{Johnson-Roberson}, ``Bio-{{LSTM}}: {{A
  Biomechanically Inspired Recurrent Neural Network}} for 3-{{D Pedestrian
  Pose}} and {{Gait Prediction}},'' \emph{IEEE Robotics and Automation
  Letters}, vol.~4, no.~2, pp. 1501--1508, Apr. 2019.

\bibitem{sangHumanMotionPrediction2020}
H.-F. Sang, Z.-Z. Chen, and D.-K. He, ``Human {{Motion}} prediction based on
  attention mechanism,'' \emph{Multimedia Tools and Applications}, vol.~79,
  no.~9, pp. 5529--5544, Mar. 2020.

\bibitem{martinezHumanMotionPrediction2017}
J.~Martinez, M.~J. Black, and J.~Romero, ``On {{Human Motion Prediction Using
  Recurrent Neural Networks}},'' in \emph{Proceedings of the {{IEEE
  Conference}} on {{Computer Vision}} and {{Pattern Recognition}}}, 2017, pp.
  2891--2900.

\bibitem{guoHumanMotionPrediction2019}
X.~Guo and J.~Choi, ``Human {{Motion Prediction}} via {{Learning Local
  Structure Representations}} and {{Temporal Dependencies}},''
  \emph{Proceedings of the AAAI Conference on Artificial Intelligence},
  vol.~33, no.~01, pp. 2580--2587, Jul. 2019.

\bibitem{coronaContextawareHumanMotion2020}
E.~Corona, A.~Pumarola, G.~Aleny{\`a}, and F.~{Moreno-Noguer}, ``Context-aware
  {{Human Motion Prediction}},'' Mar. 2020.

\bibitem{salzmannTrajectronDynamicallyFeasibleTrajectory2020}
T.~Salzmann, B.~Ivanovic, P.~Chakravarty, and M.~Pavone, ``Trajectron++:
  {{Dynamically-Feasible Trajectory Forecasting}} with {{Heterogeneous
  Data}},'' in \emph{Computer {{Vision}} \textendash{} {{ECCV}} 2020}, ser.
  Lecture {{Notes}} in {{Computer Science}}, A.~Vedaldi, H.~Bischof, T.~Brox,
  and J.-M. Frahm, Eds.\hskip 1em plus 0.5em minus 0.4em\relax {Cham}:
  {Springer International Publishing}, 2020, pp. 683--700.

\bibitem{li_neuromechanical_2018}
W.~Li and N.~P. Fey, ``Neuromechanical {Control} {Strategies} of
  {Frontal}-{Plane} {Angular} {Momentum} of {Human} {Upper} {Body} {During}
  {Locomotor} {Transitions},'' in \emph{2018 7th {IEEE} {International}
  {Conference} on {Biomedical} {Robotics} and {Biomechatronics} ({Biorob})},
  Aug. 2018, pp. 984--989, iSSN: 2155-1782.

\bibitem{chakravarty_gen-slam_2019}
\BIBentryALTinterwordspacing
P.~Chakravarty, P.~Narayanan, and T.~Roussel, ``{GEN}-{SLAM}: {Generative}
  {Modeling} for {Monocular} {Simultaneous} {Localization} and {Mapping},''
  \emph{arXiv:1902.02086 [cs]}, Feb. 2019, arXiv: 1902.02086. [Online].
  Available: \url{http://arxiv.org/abs/1902.02086}
\BIBentrySTDinterwordspacing

\bibitem{shewalkar_performance_2019}
\BIBentryALTinterwordspacing
A.~Shewalkar, D.~Nyavanandi, and S.~A. Ludwig,
  ``\BIBforeignlanguage{en}{Performance {Evaluation} of {Deep} {Neural}
  {Networks} {Applied} to {Speech} {Recognition}: {RNN}, {LSTM} and {GRU}},''
  \emph{\BIBforeignlanguage{en}{Journal of Artificial Intelligence and Soft
  Computing Research}}, vol.~9, no.~4, pp. 235--245, Oct. 2019. [Online].
  Available: \url{https://www.sciendo.com/article/10.2478/jaiscr-2019-0006}
\BIBentrySTDinterwordspacing

\bibitem{zhao_infovae_2018}
\BIBentryALTinterwordspacing
S.~Zhao, J.~Song, and S.~Ermon, ``{InfoVAE}: {Information} {Maximizing}
  {Variational} {Autoencoders},'' \emph{arXiv:1706.02262 [cs, stat]}, May 2018,
  arXiv: 1706.02262. [Online]. Available: \url{http://arxiv.org/abs/1706.02262}
\BIBentrySTDinterwordspacing

\bibitem{tan2020bump}
G.~R. Tan, M.~Raitor, and S.~H. Collins, ``Bump'em: an open-source,
  bump-emulation system for studying human balance and gait,'' in \emph{2020
  IEEE International Conference on Robotics and Automation (ICRA)}.\hskip 1em
  plus 0.5em minus 0.4em\relax IEEE, 2020, pp. 9093--9099.

\end{thebibliography}

\end{document}